%% file: main.tex
\documentclass[5p,times,authoryear]{elsarticle}

\usepackage[utf8]{inputenc}
\usepackage[T1]{fontenc}
\usepackage{amsmath,amssymb}
\usepackage{graphicx}
\usepackage{booktabs}
\usepackage{multirow}
\usepackage{xcolor}
\usepackage{siunitx}
\usepackage{dsfont}
\usepackage[hidelinks]{hyperref}
\usepackage{bm}

\graphicspath{{figures/}}
\journal{arXiv preprint}

\begin{document}

\begin{frontmatter}

\title{\textbf{A Geometry-Informed Neural Operator surrogate of LISFLOOD-FP for rapid pluvial--fluvial
flood forecasting: the case of Forl\`{i}, Italy}}

\author[reply]{Samuele Maria Garofalo\corref{cor1}}
\ead{sam.garofalo@reply.it}
\cortext[cor1]{Corresponding author.}

\author[reply]{Fabio Limardo}
\author[reply]{Simonetta Bodojra}
\author[nvidia]{Mohammad Amin Nabian}
\author[university Virginia]{Mehdi Taghizadeh}

\affiliation[reply]{organization={Data Reply},
            city={Turin},
            country={Italy}}
\affiliation[nvidia]{organization={NVIDIA},
            city={Santa Clara, CA},
            country={USA}}
\affiliation[university Virginia]{organization={University of Virginia}, country={USA}}

\input{sections/00_abstract}

\end{frontmatter}

\input{sections/01_introduction}
\input{sections/02_study_area_data}
\input{sections/03_methods}
\input{sections/04_results}

\input{sections/05_discussion}

\input{sections/06_conclusions}
\input{sections/07_backmatter}

\bibliographystyle{elsarticle-harv}
\bibliography{references}

\end{document}

%% file: sections/00_abstract.tex
\begin{abstract}
Physically based two-dimensional hydraulic models are too slow for real-time inundation forecasting and large
ensembles. We present a time-dependent
geometry-informed neural operator (GINO) surrogate of LISFLOOD-FP for the municipality of Forl\`{i} (Italy),
severely affected by the May 2023 Emilia-Romagna floods. The surrogate is designed for an operational setting in
which only river stage gauges and rain gauges are available. A reproducible pipeline turns public regional
observations into a physically based training set: each river inflow is driven by the first stage gauge
downstream of it through a calibrated water-surface boundary condition, and rain enters as SCS-CN effective
rainfall. The surrogate is an autoregressive GINO trained on 216 simulations of 24
observed events with scaled forcing. It predicts volume-conserving polygon-mean depths on an exposure-based
Voronoi mesh of 659 nodes, with a \SI{20}{min} step over \SI{72}{h}. The surrogate is tested on 38 independent
observed events that it never saw during training. On these events it reaches a critical success index (CSI) of
0.934 at \SI{0.05}{m} and 0.844 at \SI{0.3}{m}, a depth RMSE of \SI{0.008}{m} and an $R^2$ of 0.984 for
maximum-depth maps. On one NVIDIA A10G GPU a 72-h forecast
takes 1.7~s instead of 55~s with LISFLOOD-FP, and a 50-member ensemble, computed in a single batched rollout,
takes 6.2~s instead of about 46~min, which makes probabilistic flood forecasting practical.
These results show that neural-operator surrogates can deliver accurate pluvial--fluvial flood forecasts driven
by river stage for operational early warning.
\end{abstract}

\begin{keyword}
\textbf{Flood forecasting} \sep
\textbf{Geometry-informed neural operator} \sep
\textbf{Surrogate modelling} \sep
\textbf{LISFLOOD-FP} \sep
\textbf{Pluvial--fluvial flooding} \sep
\textbf{Ensemble forecasting}
\end{keyword}

%% file: sections/01_introduction.tex
\section{Introduction}\label{sec:intro}

Floods are among the most damaging natural hazards in Europe. In Italy, Emilia-Romagna is one of the most
exposed regions. In May 2023 two sequences of persistent, intense rainfall struck the Romagna foothills within a
few weeks. Many rivers overtopped or breached their embankments and flooded towns and farmland across the plain
\citep{event2023emiliaromagna}. Forl\`{i}, crossed by the Montone and Ronco rivers and located just downstream of
their Apennine catchments, was among the most affected cities. Events of this kind combine fluvial flooding from
the rivers with pluvial flooding from local rain on the urban area and the plain. They develop over one to three
days, and their timing and extent depend on how river stage and local rain evolve together. Because
both the meteorological forcing and the river response are uncertain, a single deterministic run is not enough:
an ensemble of plausible scenarios is needed.

Physically based two-dimensional hydraulic models answer these questions by solving the shallow water equations
or simplified forms of them. LISFLOOD-FP, with its local-inertial formulation \citep{bates2000lisflood,
bates2010inertial} and GPU solver \citep{shaw2021lisflood}, is one of the most efficient such codes. Even so,
running a 72-h simulation over a whole municipality for every forecast update, and for every member of an
ensemble, is still expensive. For this reason, machine-learning (ML)
surrogates that learn the response of a hydraulic model from a library of its simulations, and then reproduce
it at a small fraction of the cost, have become an active area of research \citep{bentivoglio2022review,
mosavi2018ml}.

Early surrogates relied on convolutional neural networks (CNNs) operating on regular rasters
\citep{kabir2020cnn,guo2021cnnurban,lowe2021unet}. CNNs are effective on grids, but they are tied to a fixed
structured discretisation and cannot easily represent irregular domains or meshes. Graph neural networks (GNNs)
remove this constraint by passing messages between the nodes of an unstructured mesh
\citep{pfaff2021meshgraphnets}. They have been used for spatio-temporal flood modelling with hydraulics-based
architectures \citep{bentivoglio2023swegnn} and, in HydroGraphNet, with a physics-informed loss and the
push-forward trick \citep{brandstetter2022mpnn} for stable autoregressive forecasts of pluvial and fluvial events
\citep{taghizadeh2025hydrographnet}. However, local message passing needs many layers to propagate information
across large domains. Neural operators address this by learning maps between function spaces
\citep{li2020gkn,kovachki2023neural}. The Fourier neural operator (FNO) captures long-range interactions
efficiently on regular grids \citep{li2021fno}, and FNO-type models have been applied to flood prediction
\citep{kumar2024fnoflood,xu2025neuraloperator}. The geometry-informed neural operator (GINO) combines a graph
neural operator encoder and decoder, which handle arbitrary point clouds or meshes, with an FNO on a regular
latent grid \citep{li2023gino}. FloodForecaster \citep{taghizadeh2026floodforecaster} introduced the first
time-dependent, autoregressive GINO for flood simulation. 
Operational flood forecasting in a city such as
Forl\`{i} (\SI{228.36}{km^2}) raises further questions:
\begin{itemize}
  \item \emph{Multiple rivers.} The presence of several rivers (Montone, Rabbi and Bidente) makes modeling a surrogate more complex.
  \item \emph{No discharge data.} The regional network measures river \emph{stage} and rain in near real time.
  Discharge would require rating curves, which are uncertain and often unavailable at the gauges that matter. A
  surrogate for operational use should be driven by the quantities that are actually observed.
  \item \emph{Mixed pluvial--fluvial forcing.} River inflows and local rain act together and are correlated in
  real events. The surrogate must respond to both.
  \item \emph{Product-oriented meshes.} Risk products are often delivered on administrative or exposure-based
  units rather than on the hydraulic grid. A surrogate that predicts directly on these units must preserve the
  flood volume when it aggregates the simulator output.
\end{itemize}

In this paper we adopt the time-dependent GINO model family implemented with the \texttt{neuraloperator} library
\citep{kossaifi2024neuralop}. We extend them to an operational setting driven only by stage and rain gauges and
test the surrogate on observed events that it has never seen. The main contributions are:
\begin{enumerate}
  \item A complete, reproducible pipeline from public regional observations (ARPAE stage and rain gauges,
  \citealp{arpae_dext3r}; no discharge data) to a physically based LISFLOOD-FP training set. The pipeline
  includes a first-downstream-gauge rule for driving each river inflow and a library-anchored calibration of the
  stage boundary conditions (Sections~\ref{sec:observations}--\ref{sec:dataset}).
  \item A time-dependent GINO surrogate for mixed pluvial--fluvial forcing over a whole municipality. It is trained on 24 observed events with scaled forcing. It reaches CSI of $0.934$ at \SI{0.05}{m}, $R^2$ of the maximum depth of
  $0.984$ on 38 independent observed events never seen in training
  (Section~\ref{sec:res-ff}).
  \item Volume-conserving polygon-mean targets on a product-oriented, exposure-based Voronoi mesh of 659 nodes,
  which keep the product's spatial units and leave the inference cost unchanged, combined with a loss weight on
  predicted wet nodes and a long push-forward curriculum for stable \SI{72}{h} rollouts
  (Sections~\ref{sec:mesh} and~\ref{sec:training}).
  \item A check of the simulator's credibility: the LISFLOOD-FP set-up is compared with the official perimeter of
  the areas flooded in May 2023 \citep{rer2025perimetrazione} and reaches a CSI of 0.6
  (Section~\ref{sec:res-simulator}).
  \item An evaluation that complements depth and extent scores with node-level flood metrics on the anomaly
  above each node's base level, and a 50-member ensemble for probabilistic forecasting
  (Sections~\ref{sec:res-node} and~\ref{sec:res-ensemble}).
\end{enumerate}

The paper is organised as follows. Section~\ref{sec:data} describes the study area, the observations and the
LISFLOOD-FP set-up. Section~\ref{sec:methods} presents the event dataset, the mesh, the GINO surrogate, its
training and the evaluation metrics. Section~\ref{sec:results} reports the results, from the check of the
simulator to the ensemble forecasts. Section~\ref{sec:discussion} discusses their implications and limitations,
and Section~\ref{sec:conclusions} concludes.

%% file: sections/02_study_area_data.tex
\section{Study area and data}\label{sec:data}

\subsection{Study area}\label{sec:studyarea}

\begin{figure}[t]
  \centering
  \includegraphics[width=\columnwidth]{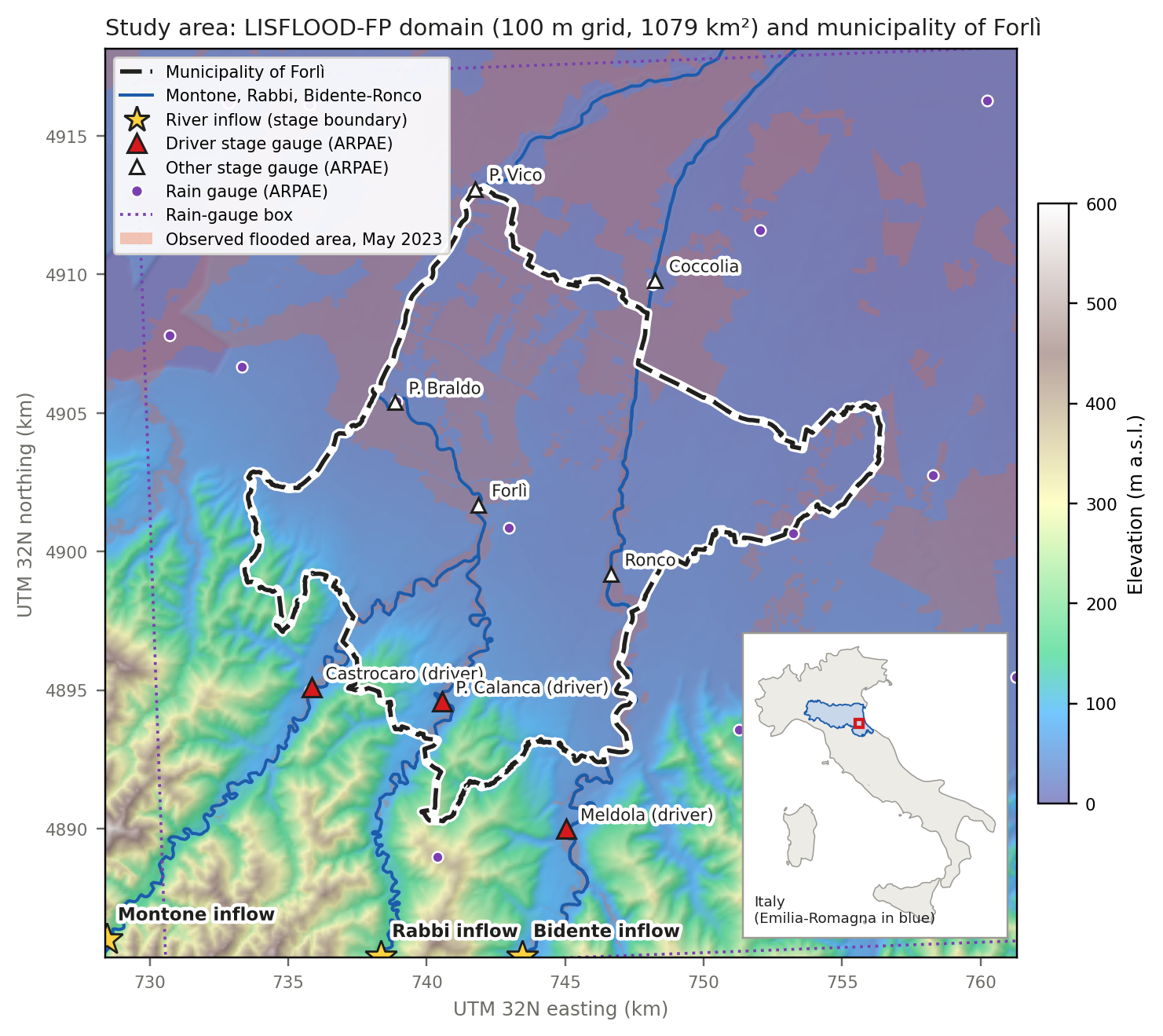}
  \caption{Study area: LISFLOOD-FP domain on the conditioned DEM, with the municipality of Forl\`i (dashed; extent of the GINO mesh, \SI{228.36}{km^2}), the Montone, Rabbi and Bidente--Ronco rivers, the three river inflow cells (stars), the ARPAE stage gauges (red: driver gauges of the three inflows), the 19 rain gauges inside the rain box (dotted) and, shaded, the areas flooded on 16--17 May 2023 \citep{rer2025perimetrazione}. Inset: location in Italy, with Emilia-Romagna in blue.}
  \label{fig:studyarea}
\end{figure}

The study area is the municipality of Forl\`i (Emilia-Romagna, northern Italy), located where the rivers draining the Romagna Apennines enter the Po plain (Fig.~\ref{fig:studyarea}). Three rivers enter the simulation domain from the south-west and south: the Montone, its tributary the Rabbi, and the Bidente, which is named Ronco downstream of Meldola, about \SI{2.6}{km} inside the domain, and flows east of the city. Downstream of the Apennine foothills the rivers are confined between levees whose crests stand several metres above the surrounding plain, so that flooding is controlled both by the river stage and by local rainfall that cannot drain to the embanked channels. The surrogate is built for the whole municipality, an area of \SI{228.36}{km^2} discretised by the mesh of Section~\ref{sec:mesh}; the LISFLOOD-FP domain extends beyond it to include the river inflows and the outflow edges.

Forl\`i was among the municipalities most severely hit by the floods of May 2023 in Emilia-Romagna \citep{event2023emiliaromagna,event2023emiliaromagnab}. On 16--17 May 2023, after a first event at the beginning of the month, persistent rainfall over already wet soils led to river overtopping and levee breaches and to the inundation of urban districts and of large parts of the plain around the city. This event provides the observed flood extent used to check the simulator (Section~\ref{sec:obsvalidation}) and is one of the base events of the training library (Section~\ref{sec:dataset}).

\subsection{Observations and geospatial data}\label{sec:observations}

\paragraph{Hydro-meteorological observations}
All forcing data are public observations of the regional environmental agency ARPAE, retrieved from its Dext3r/open-data archive \citep{arpae_dext3r}: river stage (metres above the gauge zero) and rainfall, for the years 2006--2026. No discharge measurements or rating curves are used. Each river inflow is driven by the stage of the \emph{first} ARPAE gauge downstream of the inflow, before any confluence: Castrocaro on the Montone, Ponte Calanca on the Rabbi and Meldola on the Bidente--Ronco. For each driver gauge ARPAE defines three alert thresholds $H_1<H_2<H_3$ (Castrocaro 1.3/2.0/\SI{3.0}{m}; Ponte Calanca 0.9/1.3/\SI{2.0}{m}; Meldola 0.6/1.1/\SI{2.1}{m}), where $H_3$ approximates bankfull conditions; they are used to grade event severity and to normalise the stage rise (Section~\ref{sec:lisflood}). Two upstream gauges, Rocca San Casciano (Montone) and S.~Zeno (Rabbi), serve only as substitutes to fill the shape of long gaps in the driver records. Rainfall is taken from all ARPAE rain gauges within the box \ang{44.08}--\ang{44.37}\,N, \ang{11.88}--\ang{12.30}\,E and averaged into a domain-mean hourly series. Further stage gauges inside the domain (Forl\`i and Ponte Braldo on the Montone, Ronco and Coccolia on the Ronco) are not used as forcing.

\paragraph{Terrain, river network and land cover}
The \SI{100}{m} model grid is defined on the national bare-earth DTM TINITALY (\SI{10}{m}, version~1.1) \citep{tarquini2007tinitaly,tarquini2023tinitaly}, which also fills the cells without LiDAR coverage and provides the upstream drainage areas. Elevations and channel geometry are derived from the LiDAR DTM 2023/24 of Regione Emilia-Romagna (\SI{0.5}{m} cells, nominal vertical accuracy \SI{8}{cm}, accessed at \SI{2}{m}) \citep{rer2024dtm}. River centrelines are taken from the regional topographic database (DBTR) \citep{rer_dbtr}. Land cover is the ESA WorldCover 2021 map at \SI{10}{m} \citep{zanaga2022worldcover}, used both for Manning's roughness and for the SCS curve numbers. All fine-resolution products are used only to derive \SI{100}{m} values (Section~\ref{sec:lisflood}).

\paragraph{Observed flood extent}
To check the simulator against reality we use the regional perimeter of the areas flooded on 16--17 May 2023 \citep{rer2025perimetrazione}, produced by the regional agency for territorial safety and civil protection. It was obtained by photo-interpretation of several sources, including contributions of local authorities and land-reclamation consortia, at a nominal scale of 1:5000, and maps the flooding caused by river breaches and overtopping. 

%% file: sections/03_methods.tex
\section{Methodology}\label{sec:methods}

\begin{figure*}[t]
  \centering
  \includegraphics[width=\textwidth]{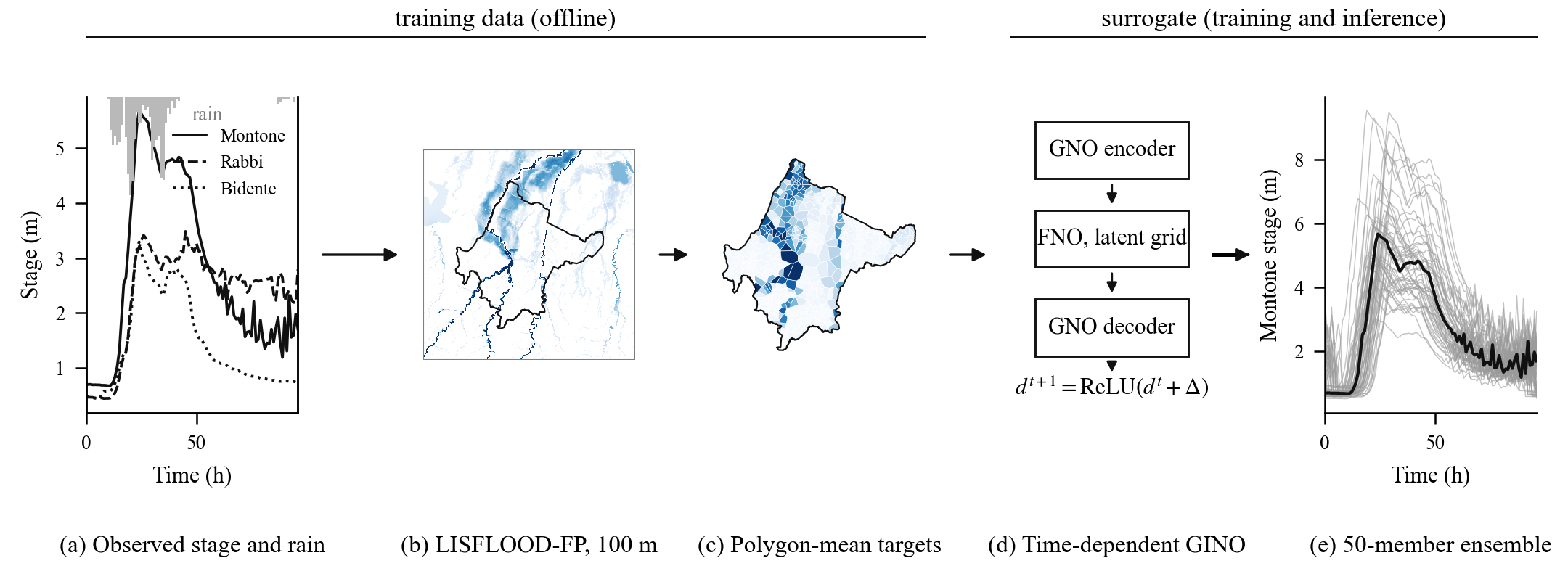}
  \caption{Overview of the workflow, illustrated with the 16 May 2023 event. (a) Observed stage at the three driver gauges and domain-mean rain (bars, inverted axis), which define the stage boundary conditions and the effective rain. (b) LISFLOOD-FP maximum water depth on the \SI{100}{m} grid. (c) Polygon-mean maximum depth on the 659-node Voronoi mesh, the training target of the surrogate. (d) Time-dependent GINO, applied autoregressively. (e) Ensemble forcing: 50 perturbed Montone stage series (grey) around the observed one (black), propagated by the surrogate in a single batched rollout.}
  \label{fig:pipeline}
\end{figure*}

The methodology chains four components (Fig.~\ref{fig:pipeline}): (i) a LISFLOOD-FP set-up of the domain driven only by observed river stage and rain (Section~\ref{sec:lisflood}); (ii) a library of historical events and a training set built scaling known hydrographs and hyetograph, with an independent set of observed test events (Sections~\ref{sec:events}--\ref{sec:dataset}); (iii) a product-oriented Voronoi mesh with volume-conserving node targets (Section~\ref{sec:mesh}); and (iv) a time-dependent GINO trained to emulate the simulator autoregressively (Sections~\ref{sec:gino}--\ref{sec:training}). The surrogate is evaluated with depth, extent and node-level flood metrics (Section~\ref{sec:metrics}) and in ensemble mode (Section~\ref{sec:ensemble}); the simulator itself is checked against the observed May 2023 flood (Section~\ref{sec:obsvalidation}).

\subsection{LISFLOOD-FP set-up}\label{sec:lisflood}

\paragraph{Solver and grid}
The training data are generated with LISFLOOD-FP~8.2 \citep{bates2000lisflood,shaw2021lisflood}, using its GPU implementation of the local-inertial (acceleration) formulation of the shallow-water equations \citep{bates2010inertial} on a regular \SI{100}{m} grid. The grid covers the bounding box of the municipality plus a \SI{5}{km} buffer: $329\times328$ cells (\num{107912} cells, about \SI{32.9}{km} $\times$ \SI{32.8}{km}, \SI{1079}{km^2}), of which the municipality covers \SI{228}{km^2}. The resolution was deliberately kept at \SI{100}{m} so that several hundred event simulations remain affordable. The GPU solver has neither subgrid channels nor infiltration, which motivates the channel conditioning and the effective-rain formulation described below. Water depth is saved every \SI{20}{min} over a \SI{72}{h} horizon, giving 217 depth frames ($t=0,\dots,216$) per run.

\paragraph{DEM conditioning}
Cell elevations are the means of the LiDAR DTM over each \SI{100}{m} cell, with the national DTM where LiDAR is missing, followed by a D8 priority-flood depression filling. Because rivers exist in the model only as DEM cells, the channel and levees are imposed explicitly: for every channel cell of the DBTR network, three LiDAR cross-sections perpendicular to the river give the bed, the bank tops (levee crests where a narrow ridge stands above the land behind) and the channel width $W$. The channel cell is set to the lowest bank top minus an equivalent depth $d_\mathrm{eq}=(z_\mathrm{bank}-z_\mathrm{bed}) \cdot \min(1,W/\Delta x)^{0.6}$, $\Delta x=\SI{100}{m}$, which gives the \SI{100}{m}-wide cell the same Manning conveyance as the real $W$-wide channel; cells flanking a levee are raised to its crest so that levees keep acting as barriers. Since LISFLOOD-FP exchanges water only across cell faces, a D4 conditioning step finally carves a face-connected path from each inflow to the domain edge along the channel, opens diagonal corners and fills residual pits, while protecting channel and levee cells. One inflow cell per river is placed on the valley bottom where the river enters the domain; all other edges are free outflow boundaries.

\paragraph{Roughness}
Manning's $n$ is assigned from the ESA WorldCover classes \citep{zanaga2022worldcover} resampled to the grid: tree cover 0.07, shrubland 0.06, grassland 0.035, cropland 0.050, built-up 0.100, bare/sparse vegetation 0.030, snow and ice 0.035, permanent water 0.025, herbaceous wetland 0.040, mangroves 0.080, moss and lichen 0.070 (unclassified cells 0.035).

\paragraph{Stage boundary conditions (HVAR)}
Without discharge data, each inflow cell receives a time-varying water-surface elevation (HVAR, Dirichlet) derived from the stage $H(t)$ of its driver gauge (Section~\ref{sec:observations}):
\begin{equation}
  \begin{aligned}
    \mathrm{WSE}(t) &= z + f\left[h_0 + s\, d_\mathrm{bf}\,\frac{H(t)-H_b}{H_3-H_b}\right],\\
    f &= \left(\frac{W}{\SI{100}{m}}\right)^{0.6},
  \end{aligned}
  \label{eq:hvar}
\end{equation}
bounded below by $z$, where $z$ is the elevation of the inflow cell, $h_0=\SI{0.2}{m}$ a low-flow depth, $d_\mathrm{bf}$ and $W$ the bankfull depth and width of the channel at the inflow (from the LiDAR sections), $H_3$ the third ARPAE threshold of the gauge (about bankfull), $H_b$ the per-event base level, taken as the 5th percentile of the stage within $\pm 30$ days of the event, and $s$ a per-river scale. The stage rise is thus applied as a fraction of bankfull, the scale on which the model channel is built, rather than in raw metres, which depend on each gauge's cross-section; the width factor $f$ maps the depth of a $W$-wide channel to the equivalent depth of a \SI{100}{m} cell. Stage records are despiked, smoothed over \SI{60}{min} and interpolated to \SI{20}{min}; each series starts from the spin-up level $z+f h_0$ and ramps linearly to the event values over the first \SI{3}{h}.

\paragraph{Calibration}
The scale $s$ of each river is calibrated once on four anchor events of the library with different severities (16 May 2023, 13 May 2019, 2 February 2018, 20 November 2024). For each anchor, river-only runs (observed stages, no rain) are performed with all scales multiplied by $m\in\{0.5,0.7,1.0,1.4,2.0\}$; for each river, the multiplier at which the simulated stage rise at the driver gauge equals the observed rise above $H_b$ is interpolated in log--log space, and the new scale is the previous one times the median over the anchors. This yields $s=1.196$ (Montone), $1.04$ (Rabbi) and $0.833$ (Bidente). The per-anchor multipliers spread by no more than about $\pm10\%$, which supports a single stage-to-inflow mapping per river across event magnitudes.

\paragraph{Effective rainfall}
As the GPU solver ignores infiltration, the rainfall input is the SCS curve-number runoff \citep{usda1986tr55} of each cell. With $P(t)$ the observed domain-mean rain cumulated from the start of the event,
\begin{equation}
  \begin{aligned}
    S &= \frac{25400}{\mathrm{CN}}-254, \qquad I_a = 0.2\,S,\\
    Q(t) &= \frac{\left(P(t)-I_a\right)^2}{P(t)-I_a+S} \quad \text{for } P(t)>I_a,
  \end{aligned}
  \label{eq:scs}
\end{equation}
($S$, $I_a$, $P$, $Q$ in mm; $Q=0$ otherwise), and the effective rain of each \SI{20}{min} step is the increment of $Q$. The curve number (AMC~II) is assigned per cell from WorldCover and the TR-55 tables for hydrologic soil group~C, in the absence of a soil map, and converted to the event's antecedent moisture class, determined from the 5-day antecedent rain with the TR-55 seasonal thresholds. The effective rain is written as a spatially distributed NetCDF forcing; rainfall itself is spatially uniform (domain mean of the gauges).

\paragraph{Spin-up}
A single spin-up run of \SI{240}{h}, without rain and with every inflow held at its low-flow level, establishes base flow in the channels. Its final state is the initial condition of every event simulation, and the same state initialises the surrogate (Section~\ref{sec:gino}).

\subsection{Event library}\label{sec:events}

Historical events are extracted from the ARPAE archives for 2006--2026, restricted to January--May and September--December. Events are the peaks of a flood index $I(t)=\max_r H_r(t)/H_{1,r}$ over the three rivers, with $I\geq 1$ and at least \SI{72}{h} between peaks; severity is 3 if a gauge exceeds its threshold $H_3$, 2 if it exceeds $H_2$, and 1 otherwise. Base events comprise all severity-3 events plus the most diverse severity-2 and severity-1 events, chosen by greedy farthest-point sampling on the per-river peak/$H_3$ ratios and the logarithm of the event rain, so that different dominant rivers and rain amounts are represented. All remaining severity~$\geq 2$ events, and a set of severity-1 events spread over the years, form the test set.

\subsection{Dataset design}\label{sec:dataset}

For the dataset construction, a limited number of base events is expanded by random scaling, and the resulting simulations are split at random into training and validation sets. Each of the 24 base events is simulated unscaled and in 8 variants in which the river-stage rise above $H_b$ and the rain are multiplied by independent factors drawn from $\mathcal{U}(0.8,1.2)$. The resulting 216 simulations are split at random into 194 training and 22 validation simulations. 

We additionally test on 38 independent observed events, simulated once with their observed forcing, never seen in training either as such or as variants. They span 2006-03-05 to 2026-02-04 and include the two largest floods of the test set, 2019-11-17 and 2024-09-18.

\subsection{Voronoi mesh and volume-conserving targets}\label{sec:mesh}

\begin{figure}[t]
  \centering
  \includegraphics[width=\columnwidth]{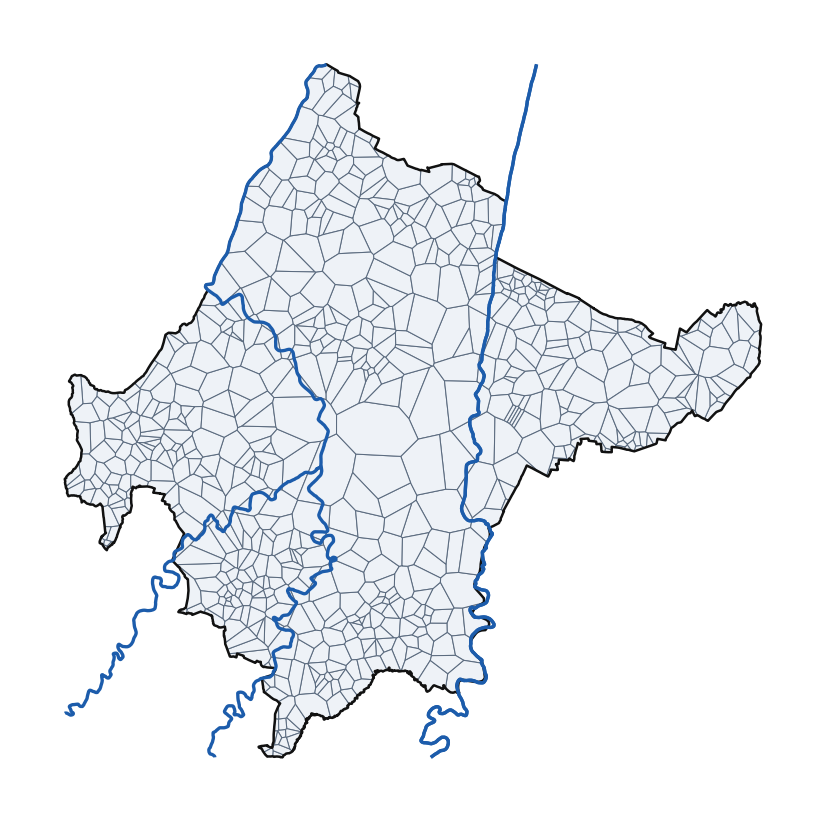}
  \caption{Exposure-based Voronoi mesh of the municipality of Forl\`i (659 polygons; median area \SI{25}{ha}), with the main rivers.}
  \label{fig:mesh}
\end{figure}

The surrogate operates on an unstructured mesh of 659 Voronoi polygons covering the municipality (Fig.~\ref{fig:mesh}). The seeds are placed according to exposure, i.e., counts of buildings, transport infrastructure, geological and water features, so that the mesh is finer where assets are concentrated. The same polygons are the aggregation unit of the risk product, so the surrogate predicts directly at the scale at which flood information is consumed.

The target depth of a node is the \emph{mean} depth over the \SI{100}{m} LISFLOOD-FP cells whose centres fall inside its polygon, i.e., the polygon water volume divided by its area; polygons smaller than a cell take the cell containing their centroid. This aggregation conserves the simulated flood volume. For consistency, the static features Elevation, Slope, Curvature and Manning are also polygon means of the DEM and roughness map used by the simulations. The aggregation is performed once during data preparation and leaves the inference cost of the surrogate unchanged.

\subsection{Time-dependent GINO}\label{sec:gino}

\begin{figure*}[t]
  \centering
  \includegraphics[width=\textwidth]{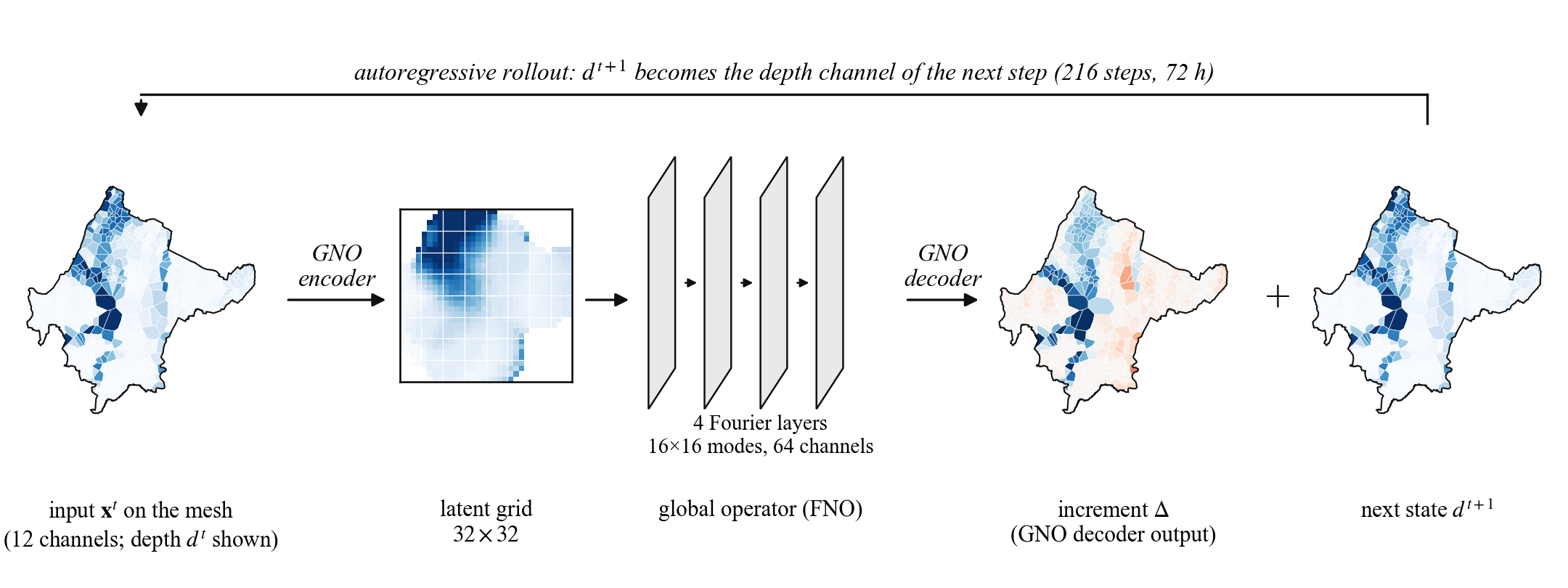}
  \caption{Architecture of the time-dependent GINO surrogate. The node features on the Voronoi mesh are encoded onto a regular $32\times32$ latent grid by a GNO, processed globally by four Fourier layers and decoded back to the nodes by a second GNO, which predicts the depth increment of Eq.~\eqref{eq:residual}; the new depth is fed back as input for the next \SI{20}{min} step.}
  \label{fig:architecture}
\end{figure*}

We adopt the geometry-informed neural operator \citep{li2023gino} as implemented in the \texttt{neuraloperator} library, version~2.0.0 \citep{kossaifi2024neuralop,neuraloperator_software}, in the time-dependent autoregressive formulation of \citet{taghizadeh2026floodforecaster} (Fig.~\ref{fig:architecture}). A graph-neural-operator (GNO) encoder \citep{li2020gkn} maps the node features from the irregular mesh to a regular $32\times32$ latent grid, by kernel integration over all mesh nodes within a radius of 0.15 in coordinates normalised to the unit square; a Fourier neural operator \citep{li2021fno,kovachki2023neural} with $16\times16$ retained modes propagates information globally on the latent grid; and a GNO decoder with the same radius maps the latent field back to the mesh nodes. Other architectural hyper-parameters are the defaults of \texttt{neuraloperator} version~2.0.0 (four FNO layers with 64 hidden channels; Table~\ref{tab:hyper}), for a total of about 2.7 million trainable parameters.

At each step, the model $\mathcal{G}_\theta$ receives the input $\mathbf{x}^t$ (Section~\ref{sec:inputs}) and predicts the normalised depth increment, which is added to the current state:
\begin{equation}
  \mathbf{d}^{t+1} = \mathrm{ReLU}\!\left(\mathbf{d}^{t} + \sigma_\Delta\,\mathcal{G}_\theta(\mathbf{x}^t) + \mu_\Delta\right),
  \qquad t=0,\dots,215,
  \label{eq:residual}
\end{equation}
where $\mathbf{d}^t$ is the vector of node depths, $\mu_\Delta$ and $\sigma_\Delta$ are the mean and standard deviation of the one-step increments over the training set, and the ReLU enforces non-negative depths. The time step is \SI{20}{min}, and a \SI{72}{h} forecast requires 216 steps.

\subsubsection{Input features}\label{sec:inputs}

Each node carries 12 input channels (Table~\ref{tab:inputs}). The six static channels describe the terrain, the roughness, the polygon area and the proximity to the river inflows. The five dynamic channels encode the forcing at the current step: rain enters as the domain-mean effective rain of Eq.~\eqref{eq:scs} together with its cumulative value and a short causal moving mean; the river forcing enters through the HVAR water-surface elevations of the three inflow points, each projected onto its nearest mesh node (anchor) and spread to the other nodes with a Gaussian kernel ($\sigma=\SI{1000}{m}$), the rivers being combined by a kernel-weighted mean. The resulting field is expressed as an anomaly above a per-node reference level, the minimum over the training simulations, so that it is zero at low flow; a \SI{6}{h} causal moving mean adds memory of the hydrograph. The last channel is the current depth. Static and rain channels are standardised, and the river channels are only rescaled so that zero remains zero; all normalisation statistics are computed on the training simulations.

\begin{table}[t]
  \centering
  \small
  \caption{Input channels of the GINO surrogate (per node and time step).}
  \label{tab:inputs}
  \begin{tabular}{p{0.20\columnwidth}p{0.70\columnwidth}}
    \toprule
    Channel & Description \\
    \midrule
    \multicolumn{2}{l}{\emph{Static}} \\
    Elevation & Polygon mean of the conditioned DEM (m) \\
    Slope & Polygon mean of the terrain slope \\
    Curvature & Polygon mean of the terrain curvature \\
    Manning & Polygon mean of Manning's $n$ \\
    Area & Polygon area \\
    SDF & $\exp(-d/\SI{1000}{m})$, $d$ = distance to the nearest inflow anchor node \\
    \midrule
    \multicolumn{2}{l}{\emph{Dynamic (forcing)}} \\
    Precip & Domain-mean effective (SCS-CN) rain at $t$ \\
    Rain\_cum & Effective rain cumulated up to $t$ \\
    Rain\_win & Causal mean over 6 steps (\SI{2}{h}) \\
    BC\_anom & Kernel-propagated inflow WSE above the node reference level (m) \\
    BC\_win & Causal mean of BC\_anom over 18 steps (\SI{6}{h}) \\
    \midrule
    \multicolumn{2}{l}{\emph{State}} \\
    Depth & Node depth at $t$ (teacher-forced in training, fed back at inference) \\
    \bottomrule
  \end{tabular}
\end{table}

\subsection{Training}\label{sec:training}

\paragraph{Loss}
Training samples are pairs $(\mathbf{x}^t,\Delta^t)$ drawn from all steps of the 194 training simulations, with $\Delta^t$ the normalised true increment. Since most nodes are dry most of the time, a plain MSE would be dominated by trivial zeros. We therefore use a wet-weighted MSE:
\begin{equation}
  \begin{aligned}
    \mathcal{L} &= \frac{\sum_{n} w_n\,(\hat{\Delta}_n-\Delta_n)^2}{\sum_n w_n},\\
    w_n &= 1 + \lambda\,\mathds{1}\!\left[\max\!\left(d_n^{t},\,d_n^{t+1},\,\hat d_n^{t+1}\right) > h_w\right],
  \end{aligned}
  \label{eq:loss}
\end{equation}
with $\lambda=5$ and $h_w=\SI{0.05}{m}$, where $d$ is the true depth and $\hat d^{t+1}$ the depth predicted by Eq.~\eqref{eq:residual}. The predicted-wet term gives the same weight to spurious thin water produced by the model as to a missed wet node, which discourages the slow accumulation of false water during long rollouts.

\paragraph{Stabilisation of the rollout}
Two techniques counter the drift of autoregressive rollouts. First, Gaussian noise with standard deviation $0.1\,\sigma_d$ ($\sigma_d$ = standard deviation of the training depths) is added to the input depth of one-step samples, with the target increment recomputed from the noisy input. Second, a push-forward curriculum \citep{brandstetter2022mpnn}: with probability 0.3, a batch is unrolled for $K$ steps, the first $K-1$ without gradient on the model's own predictions, and the last one is penalised against the true depth at $t+K$. The unroll length grows with the epochs: $K=1$ (epochs 1--20), 2 (21--40), 8 (41--60), 16 (61--80) and 32 (81--100).

\paragraph{Optimisation and model selection}
The model is trained for 100 epochs with AdamW \citep{loshchilov2019adamw} (learning rate $10^{-4}$, weight decay $10^{-5}$, 5 epochs of linear warm-up followed by cosine annealing), batch size 16 and gradient-norm clipping at 1. Every 10 epochs the model is run in free-running mode over the full \SI{72}{h} of all validation simulations, starting from the spin-up state, and the checkpoint with the lowest area-weighted mean absolute error on wet nodes (node-level definitions of Section~\ref{sec:metrics}) is retained. Training used PyTorch~2.6 \citep{paszke2019pytorch} on an AWS SageMaker \texttt{ml.g5.2xlarge} instance with one NVIDIA A10G GPU (\SI{24}{GB}), in \SI{4.47}{h}. Hyper-parameters are summarised in Table~\ref{tab:hyper}.

\begin{table}[t]
  \centering
  \small
  \caption{Hyper-parameters of the GINO surrogate and of its training.}
  \label{tab:hyper}
  \begin{tabular}{p{0.52\columnwidth}p{0.38\columnwidth}}
    \toprule
    Parameter & Value \\
    \midrule
    Input / output channels & 12 / 1 (depth increment) \\
    GNO radius (encoder, decoder) & 0.15 (normalised coordinates) \\
    Latent grid & $32\times32$ \\
    FNO modes & $16\times16$ \\
    FNO layers / hidden channels & 4 / 64 (GELU) \\
    GNO kernel MLPs (encoder; decoder) & $[80,80,80]$; $[512,256]$, linear transform \\
    Positional embedding channels & 32 \\
    Other architecture settings & \texttt{neuraloperator} 2.0.0 defaults \\
    Trainable parameters & \num{2733629} \\
    \midrule
    Optimiser & AdamW, lr $10^{-4}$, weight decay $10^{-5}$ \\
    Schedule & 5-epoch warm-up, cosine annealing \\
    Epochs / batch size & 100 / 16 \\
    Gradient clipping & 1.0 (norm) \\
    Wet weight $\lambda$ / threshold $h_w$ & 5 / \SI{0.05}{m}, incl.\ predicted wet \\
    Input-depth noise & $0.1\,\sigma_d$ ($K=1$ batches) \\
    Push-forward probability & 0.3 \\
    Unroll schedule $K$ & 1, 2, 8, 16, 32 (20 epochs each) \\
    Rollout validation & every 10 epochs \\
    Rain\_win / BC\_win windows & 6 / 18 steps \\
    BC kernel $\sigma$ & \SI{1000}{m} \\
    \midrule
    Hardware & 1$\times$ NVIDIA A10G (\SI{24}{GB}) \\
    Training time & \SI{4.47}{h} \\
    \bottomrule
  \end{tabular}
\end{table}

\subsection{Evaluation metrics}\label{sec:metrics}

All metrics are computed on free-running rollouts of 216 steps (\SI{72}{h}) started from the spin-up state, for the 22 validation simulations (scaled variants of the training events) and the 38 test events (independent observed events).

\paragraph{Depth and extent metrics}
Following \citet{taghizadeh2026floodforecaster}, accuracy is measured on the absolute depth at all $N=659$ nodes and all time steps. For a simulation with predicted and true depths $\hat d_n^t$ and $d_n^t$,
\begin{equation}
  \mathrm{RMSE}_t = \frac{1}{T}\sum_{t=1}^{T}\sqrt{\sum_{n=1}^{N} \omega_n\left(\hat d_n^t - d_n^t\right)^2},
  \label{eq:rmse}
\end{equation}
\begin{equation}
  \mathrm{CSI}_{\mathrm{thr}} = \frac{1}{T}\sum_{t=1}^{T}\frac{H_{\mathrm{thr}}^t}{H_{\mathrm{thr}}^t + M_{\mathrm{thr}}^t + F_{\mathrm{thr}}^t},
  \label{eq:csi}
\end{equation}
where $T=216$, $\omega_n=1/N$, and $H$, $M$ and $F$ are the (weighted) numbers of hits, misses and false alarms of the wet maps $\hat d>\mathrm{thr}$ and $d>\mathrm{thr}$, for $\mathrm{thr}=0.05$ and \SI{0.3}{m}. Each simulation thus yields one value, and we report the mean $\pm$ standard deviation over simulations. Because Voronoi polygons differ in area, we also report area-weighted versions ($\omega_n = A_n/\sum_m A_m$ and $H$, $M$, $F$ summed in area). To characterise maximum values and hazard timing, we further compute, per simulation and over nodes, the coefficient of determination $R^2$ between true and predicted maximum-depth maps, between arrival times (first step with depth $>\SI{0.1}{m}$, on nodes wet in both) and between inundation durations (time with depth $>\SI{0.1}{m}$, on nodes wet in either).

\paragraph{Node-level flood metrics}
To focus on the flood signal we define, for each node, a base level $b_n$ as the minimum over the training simulations of the per-simulation maximum depth, and the anomaly
\begin{equation}
  a_n^t = \max\!\left(d_n^t - b_n,\ 0\right).
  \label{eq:anomaly}
\end{equation}
Nodes that are deeper than \SI{1}{m} even in the weakest event (permanently deep channel nodes) and isolated deep pits are excluded; a node is wet when $a_n^t>\SI{5}{cm}$, and the flood volume is $\sum_n a_n^t A_n$. Because node depths are polygon means, polygons crossed by a river also contain the in-channel stage rise, so these metrics reflect near-channel as well as overbank dynamics. For each test event we compute, at the step of the true volume peak, the area-weighted CSI and false-alarm ratio (FAR) and the ratio of predicted to true volume; the area-weighted MAE over nodes wet in the truth or in the prediction over the whole \SI{72}{h}; and the timing error of the predicted volume peak; these scores are averaged over all test events. Between-event diagnostics complement these scores: the correlation between true and predicted peak volumes, their coefficients of variation, and the one-step (teacher-forced) skill relative to persistence in the rising, peak and recession phases.

\subsection{Ensemble forecasting}\label{sec:ensemble}

Because a rollout costs a small fraction of a LISFLOOD-FP run (Section~\ref{sec:res-efficiency}), the surrogate can propagate forcing uncertainty by ensembles. For each test event we generate 50 members by perturbing the river forcing, with a lognormal amplitude factor ($\sigma=0.30$) applied to the anomaly above the event minimum and a time shift drawn from a normal distribution with standard deviation 12 steps (\SI{4}{h}), and the rain, with a lognormal amplitude factor ($\sigma=0.15$) and a time shift with standard deviation 3 steps (\SI{1}{h}). All members are run in one batched rollout. At the true peak we evaluate the ensemble spread on wet nodes, the ratio of spread to the error of the ensemble mean (1 for a consistent ensemble) and the fraction of nodes with an informative flood probability (anomaly $>\SI{10}{cm}$ with probability between 5\% and 95\%). Since the error is measured against the unperturbed LISFLOOD-FP run, the spread--error ratio is a consistency check rather than a calibration measure.

\subsection{Comparison of LISFLOOD-FP with the observed flood extent}\label{sec:obsvalidation}

The credibility of the training data is assessed by comparing the LISFLOOD-FP simulation of the 16 May 2023 event, run with the same set-up as the dataset, with the regional perimeter of the areas flooded on 16--17 May 2023 (Section~\ref{sec:observations}). The comparison protocol is as follows. (i) The simulated flood map is the maximum depth over the \SI{72}{h} horizon, thresholded at \SI{0.10}{m}, as in previous binary-pattern evaluations \citep{aronica2002,travert2024measures}. (ii) The main channel itself is excluded. (iii) The comparison is restricted to the municipality of Forl\`i and carried out on the \SI{100}{m} cells, a cell being observed as flooded when at least 30\% of its area is covered by the perimeter. Agreement is summarised by the critical success index $\mathrm{CSI}=H/(H+M+F)$ \citep{wing2017validation,stephens2014binary}, with $H$, $M$ and $F$ the numbers of hit, missed and false-alarm cells. The perimeter is an envelope of the event without timing information, so the comparison checks the simulated maximum extent only.

%% file: sections/04_results.tex
\section{Results}\label{sec:results}

\subsection{Credibility of the LISFLOOD-FP training data}\label{sec:res-simulator}

\begin{figure}[!tbp]
  \centering
  \includegraphics[width=\columnwidth]{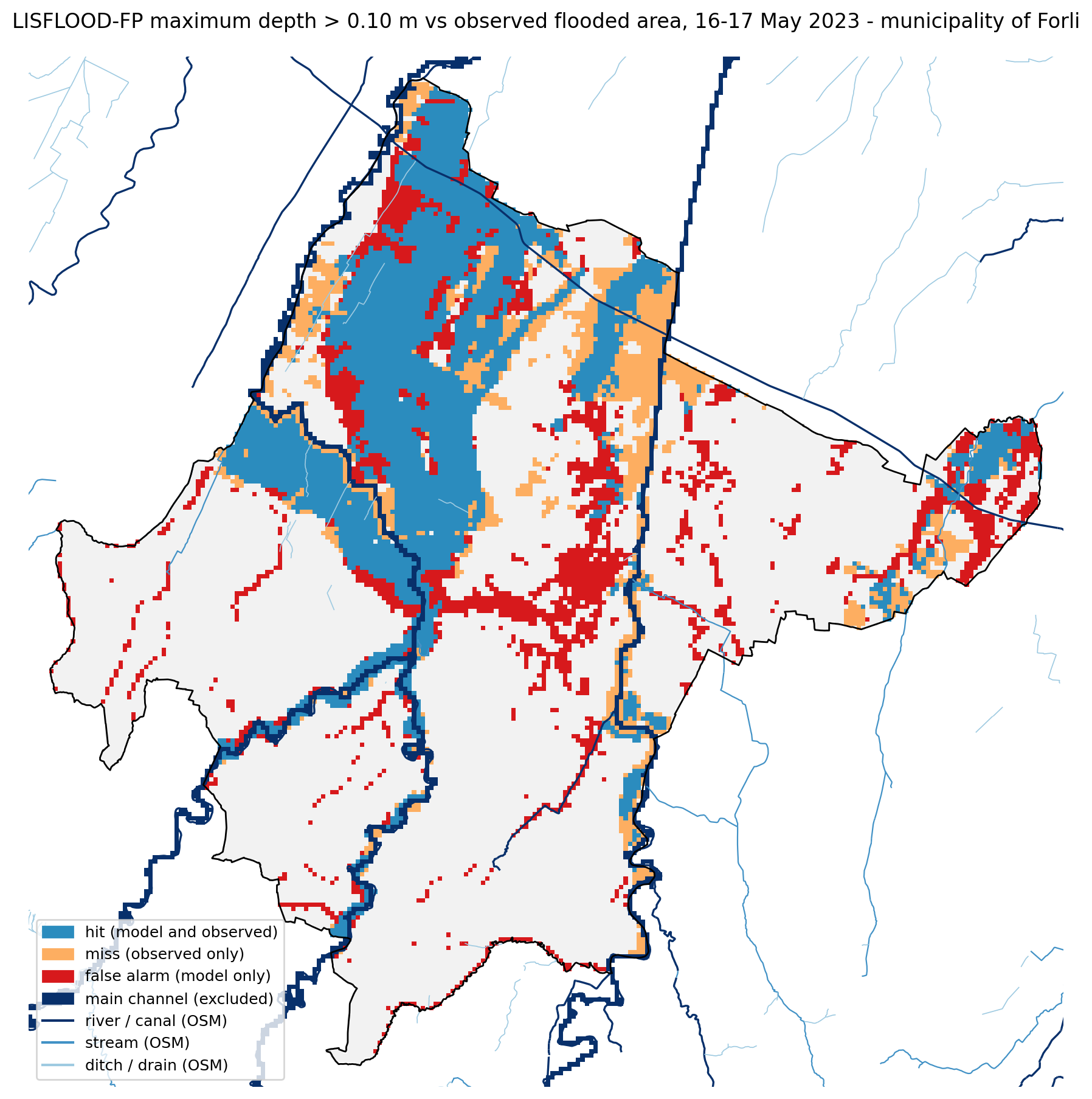}
  \caption{LISFLOOD-FP maximum water depth above \SI{0.10}{m} over \SI{72}{h} for the 16 May 2023 event compared, cell by cell, with the regional perimeter of the areas flooded on 16--17 May 2023 \citep{rer2025perimetrazione}, municipality of Forl\`i.}
  \label{fig:lisflood-obs}
\end{figure}

\begin{figure*}[!tb]
  \centering
  \includegraphics[width=\textwidth]{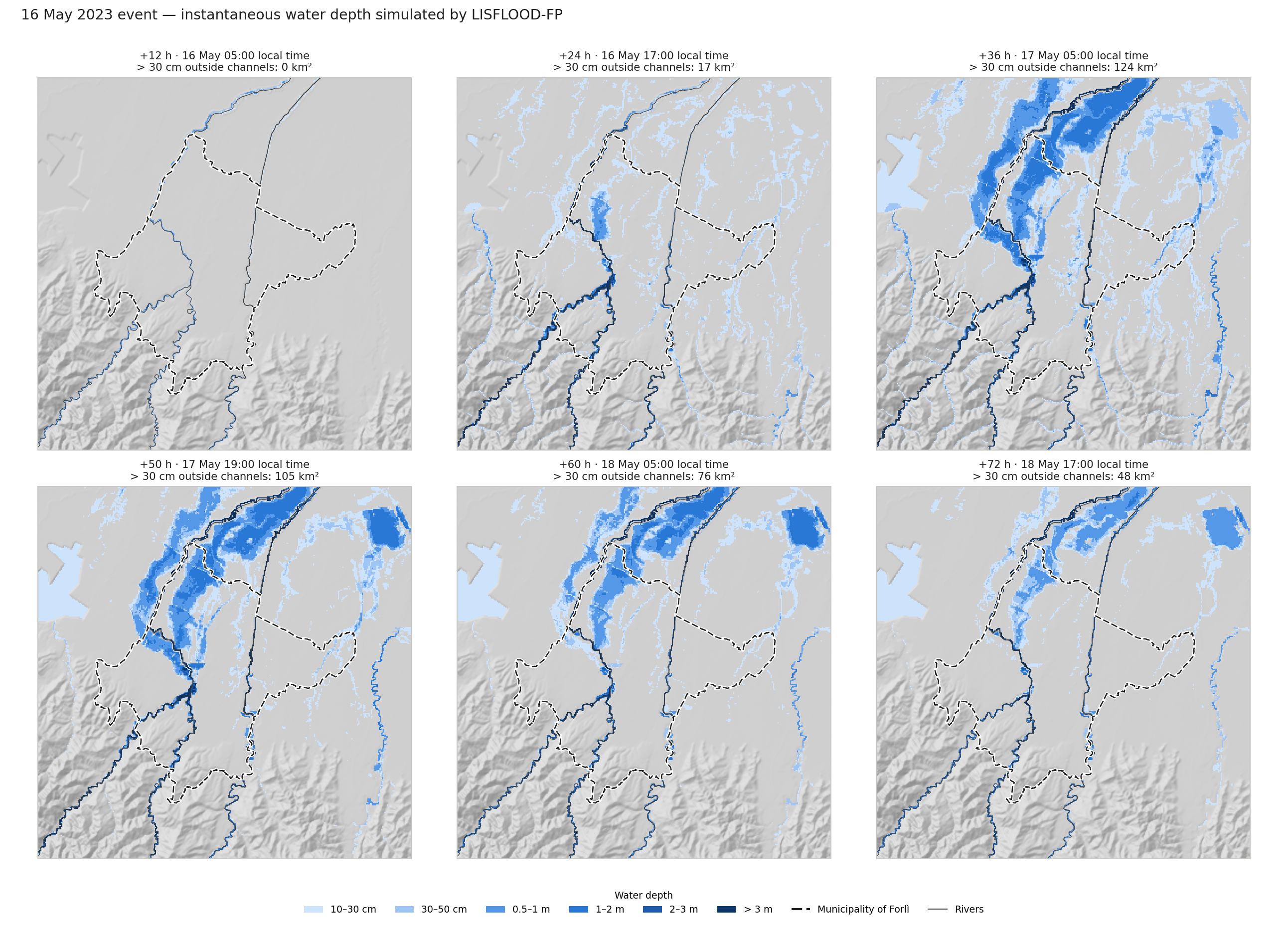}
  \caption{Water depth simulated by LISFLOOD-FP for the 16 May 2023 event, from \SI{12}{h} to \SI{72}{h} after the start of the simulation window.}
  \label{fig:sequence}
\end{figure*}

\begin{figure*}[!tb]
  \centering
  \includegraphics[width=\textwidth]{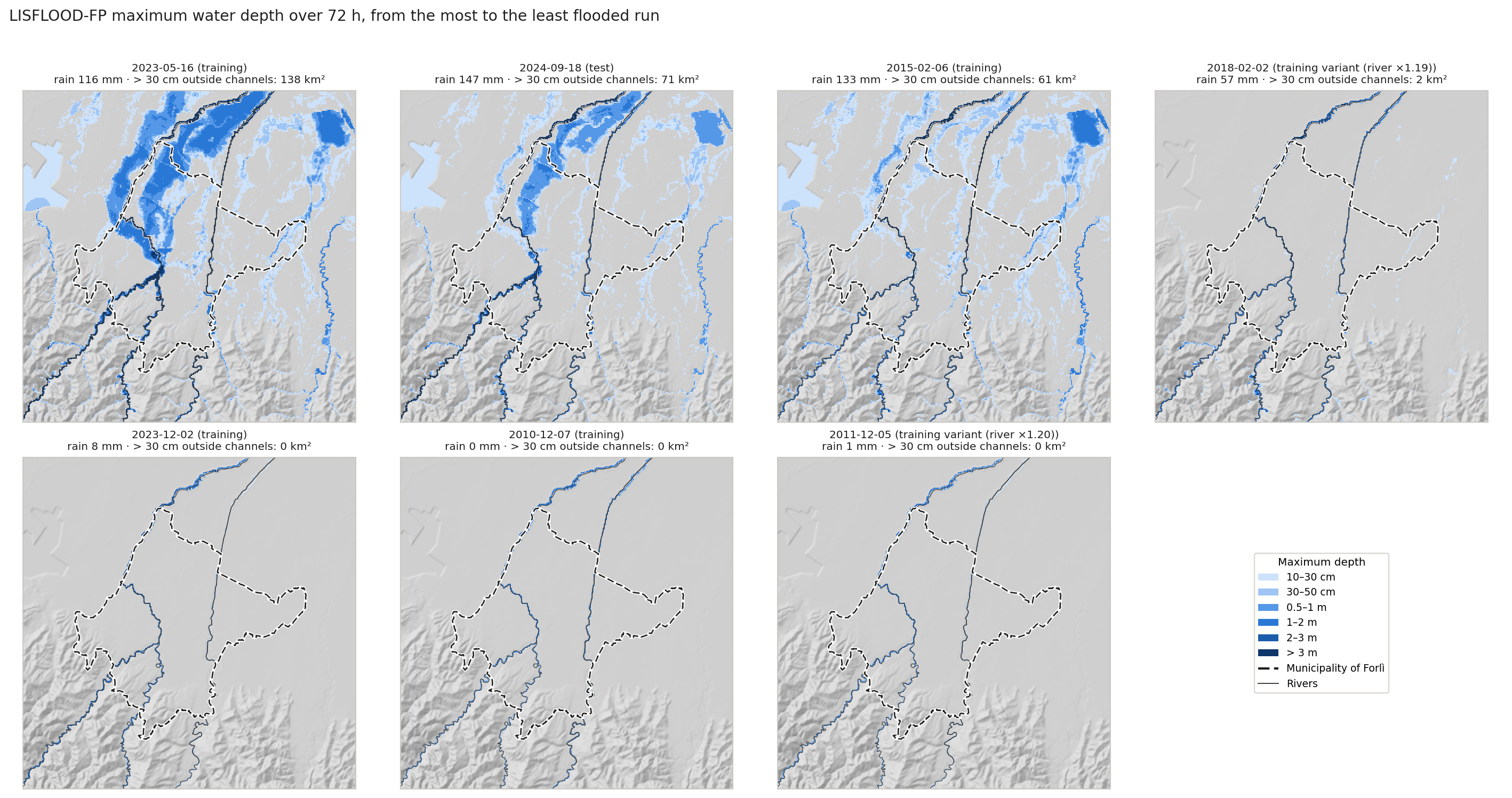}
  \caption{Maximum water depth simulated by LISFLOOD-FP over \SI{72}{h} for a selection of runs of the final set-up, ordered from the most to the least flooded.}
  \label{fig:maxdepth}
\end{figure*}

Our training set is driven by observed gauge records, which allows the simulator set-up itself to be checked against an observed flood. Following the protocol of Section~\ref{sec:obsvalidation}, the maximum water depth simulated by LISFLOOD-FP for the 16 May 2023 event was compared with the regional perimeter of the areas flooded on 16--17 May 2023 \citep{rer2025perimetrazione}, using the \SI{0.10}{m} wet threshold adopted in previous binary-pattern evaluations \citep{aronica2002,travert2024measures}. Over the municipality of Forl\`i, LISFLOOD-FP reaches a critical success index of $\mathrm{CSI}=0.6$. For a \SI{100}{m} model without subgrid channels, driven only by stage and rain gauges, this indicates a credible reproduction of the observed flood pattern, supporting the use of these simulations as training data.

Figure~\ref{fig:lisflood-obs} shows the spatial pattern behind the score. The bulk of the observed flooding in the plain north-west of the city, between the Montone and the north-western municipal boundary, is reproduced as hits over a large contiguous area. Misses concentrate along the margins of the observed polygons and in a band on the eastern side of the municipality, while many of the model-only cells are small, scattered patches of water away from the main rivers. Figure~\ref{fig:sequence} shows the evolution in time of the same flood simulated by LISFLOOD-FP.

Figure~\ref{fig:maxdepth} illustrates the range of hydraulic conditions covered by the simulations. The largest events, such as the May 2023 training event and the September 2024 test event, flood extensive areas of the plain downstream of the city, with maximum depths of 0.5--\SI{2}{m} along the Montone corridor and north of the municipality, whereas low-forcing events keep the water within the channel network. This spread, from in-channel conditions to municipality-scale inundation, is what the surrogate has to reproduce from the same set of inputs.

\subsection{Accuracy of the surrogate}\label{sec:res-ff}

\begin{table*}[!tb]
  \centering
  \small
  \caption{Accuracy of the GINO surrogate on the 22 validation simulations and on the 38 independent test events, mean $\pm$ standard deviation over simulations.}
  \label{tab:ffmetrics}
  \begin{tabular}{lcc}
    \toprule
    Metric & Validation ($n=22$) & Test ($n=38$) \\
    \midrule
    RMSE (m)                          & $0.006 \pm 0.007$ & $0.008 \pm 0.009$ \\
    CSI (\SI{0.05}{m})                & $0.936 \pm 0.045$ & $0.934 \pm 0.039$ \\
    CSI (\SI{0.3}{m})                 & $0.901 \pm 0.081$ & $0.844 \pm 0.084$ \\
    RMSE, area-weighted (m)           & $0.007 \pm 0.008$ & $0.009 \pm 0.010$ \\
    CSI (\SI{0.05}{m}), area-weighted & $0.942 \pm 0.039$ & $0.936 \pm 0.027$ \\
    CSI (\SI{0.3}{m}), area-weighted  & $0.899 \pm 0.079$ & $0.836 \pm 0.089$ \\
    $R^2$ maximum depth               & $0.988 \pm 0.014$ & $0.984 \pm 0.017$ \\
    $R^2$ arrival time                & $0.921 \pm 0.091$ & $0.856 \pm 0.219$ \\
    $R^2$ inundation duration         & $0.929 \pm 0.059$ & $0.876 \pm 0.096$ \\
    \bottomrule
  \end{tabular}
\end{table*}

Table~\ref{tab:ffmetrics} reports the depth and extent metrics, computed on absolute depth at all nodes and over all 216 autoregressive steps (\SI{72}{h}), for the 22 validation simulations and for the 38 independent test events. On the test events the GINO surrogate achieves a time-averaged depth RMSE of $0.008 \pm 0.009$\,m and a CSI of $0.934 \pm 0.039$ at \SI{0.05}{m} and $0.844 \pm 0.084$ at \SI{0.3}{m}. Maximum-depth maps are reproduced with $R^2 = 0.984 \pm 0.017$, and the hazard metrics derived from the full time series, arrival time and inundation duration, reach $R^2 = 0.856 \pm 0.219$ and $0.876 \pm 0.096$. Area-weighted scores, which give each node a weight proportional to its polygon area, are consistent with the node-averaged ones (test CSI of $0.936 \pm 0.027$ at \SI{0.05}{m} and $0.836 \pm 0.089$ at \SI{0.3}{m}).

Validation and test scores are very close (CSI at \SI{0.05}{m} of $0.936$ and $0.934$; $R^2$ of the maximum depth of $0.988$ and $0.984$). The validation simulations are scaled variants of the training events, whereas the test events are observed events with their own hydrograph shapes, rain patterns and timing, never seen in any form during training. The surrogate therefore generalises from 24 base events to new observed events with no loss of skill in the detection of inundated nodes.

\subsection{Node-level flood metrics}\label{sec:res-node}

\begin{figure*}[!tb]
  \centering
  \includegraphics[width=\textwidth]{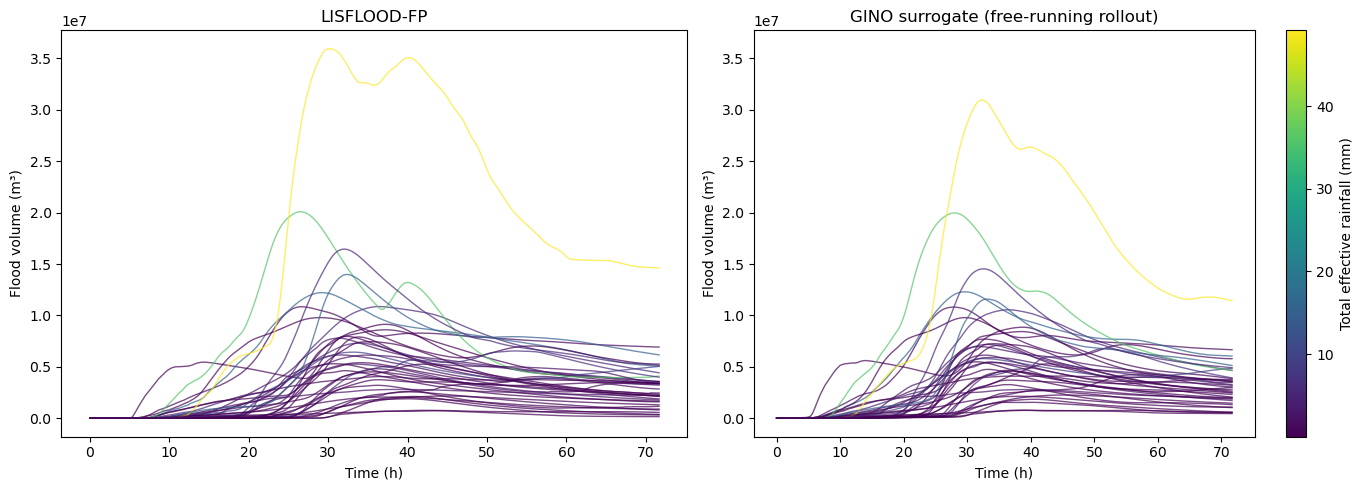}
  \caption{Flood volume (sum over nodes of the depth anomaly times the polygon area, \si{m^3}) for all 38 test events: LISFLOOD-FP (left) and free-running GINO surrogate rollout over \SI{72}{h} (right). Line colour: total effective rainfall of the event (mm).}
  \label{fig:trajectories}
\end{figure*}

\begin{figure*}[!tb]
  \centering
  \includegraphics[width=\textwidth]{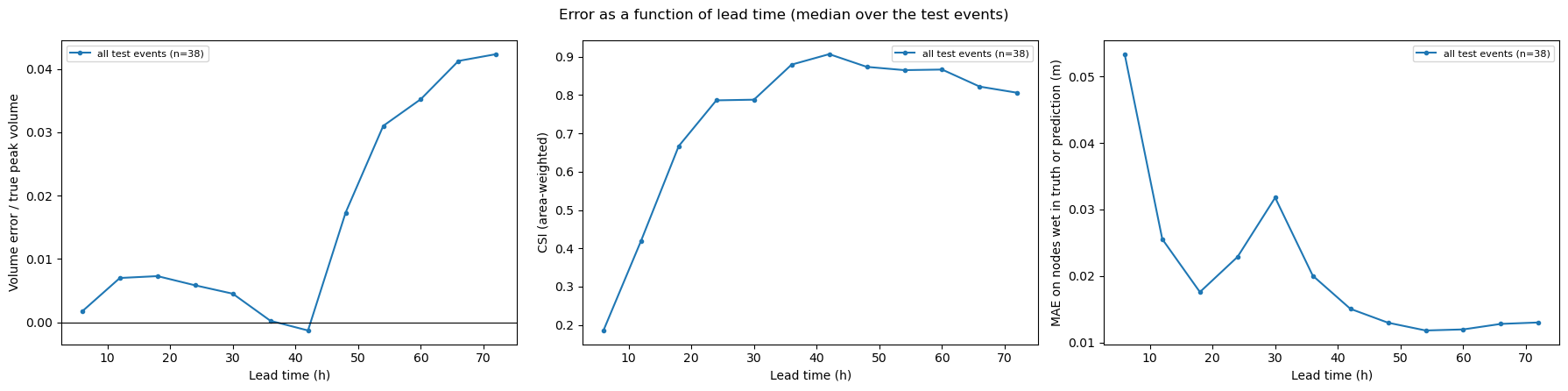}
  \caption{Error as a function of lead time, median over the 38 test events in \SI{6}{h} bins: (left) volume error relative to the true peak volume; (centre) area-weighted CSI; (right) MAE on nodes wet in truth or prediction (m). Node-level flood metrics on the anomaly above each node's base level.}
  \label{fig:leadtime}
\end{figure*}

To isolate the flood signal we also evaluate node-level flood metrics, defined on the anomaly above each node's base level (Section~\ref{sec:metrics}). Averaged over the 38 test events, the surrogate reaches an area-weighted CSI at the true volume peak of $0.840$ (90\% bootstrap confidence interval $[0.792, 0.885]$; median over events $0.904$), with a false-alarm ratio of $0.072$ ($[0.040, 0.112]$). It reproduces the peak flood volume with a mean ratio of $0.971$ ($[0.954, 0.988]$) and has a mean absolute error on wet nodes of \SI{0.020}{m} ($[0.017, 0.025]$) over the full \SI{72}{h}. The peak is predicted on average \SI{0.73}{h} late, close to two output intervals.

The surrogate also ranks events correctly by magnitude, which is essential for warning. Across test events, the correlation between true and predicted peak flood volume is $r = 0.995$. The predicted between-event variability is slightly compressed (coefficient of variation of the peak volume $0.799$ against $0.864$ in the simulations), and the dependence of the flood volume on rainfall is preserved ($r = 0.873$ predicted against $0.877$ true).

Figure~\ref{fig:trajectories} shows the free-running trajectories of the flood volume for all test events. The surrogate reproduces the timing, the shape of the rising limb and the recession of each event over the full \SI{72}{h}, and the ordering of events by magnitude. Figure~\ref{fig:leadtime} shows how the errors evolve with lead time. The median volume error stays within 1\% of the peak volume up to about \SI{42}{h} and grows to about 4\% at the end of the recession, where the surrogate retains slightly too much water; there is no sign of the compounding error growth typical of unstable autoregressive models. The CSI is low only in the first hours, when very few nodes are wet and the score is dominated by small absolute differences; it exceeds $0.8$ from about \SI{24}{h}, peaks at about $0.9$ around \SI{42}{h} and remains above $0.8$ until the end of the horizon. The MAE on wet nodes drops after the first hours, shows a temporary increase around \SI{30}{h}, when the flood is spreading fastest, and settles at about \SI{0.012}{m} during the recession.

\subsection{Representative and challenging events}\label{sec:res-cases}

\begin{figure*}[!tb]
  \centering
  \includegraphics[width=\textwidth]{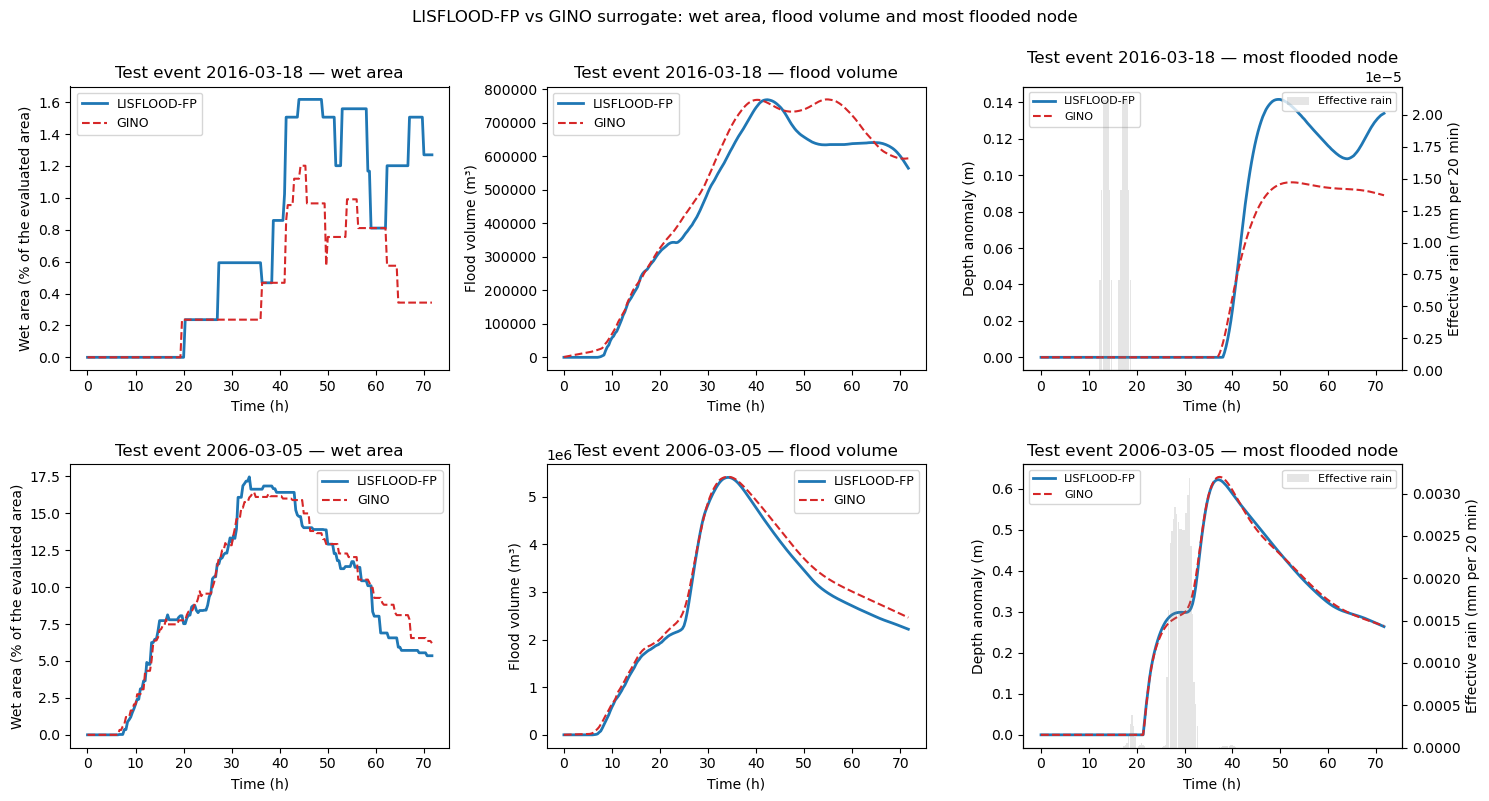}
  \caption{Two test events of different magnitude: 2016-03-18 (top) and 2006-03-05 (bottom). Left: wet fraction of the municipality (\% of the area with anomaly above \SI{5}{cm}); centre: flood volume (sum of anomaly times area, \si{m^3}); right: depth anomaly at the node with the largest anomaly, with domain-mean rainfall (grey bars, right axis). Solid blue: LISFLOOD-FP; dashed red: GINO surrogate (free-running rollout, \SI{72}{h}).}
  \label{fig:overview}
\end{figure*}

\begin{figure*}[!tb]
  \centering
  \includegraphics[width=\textwidth]{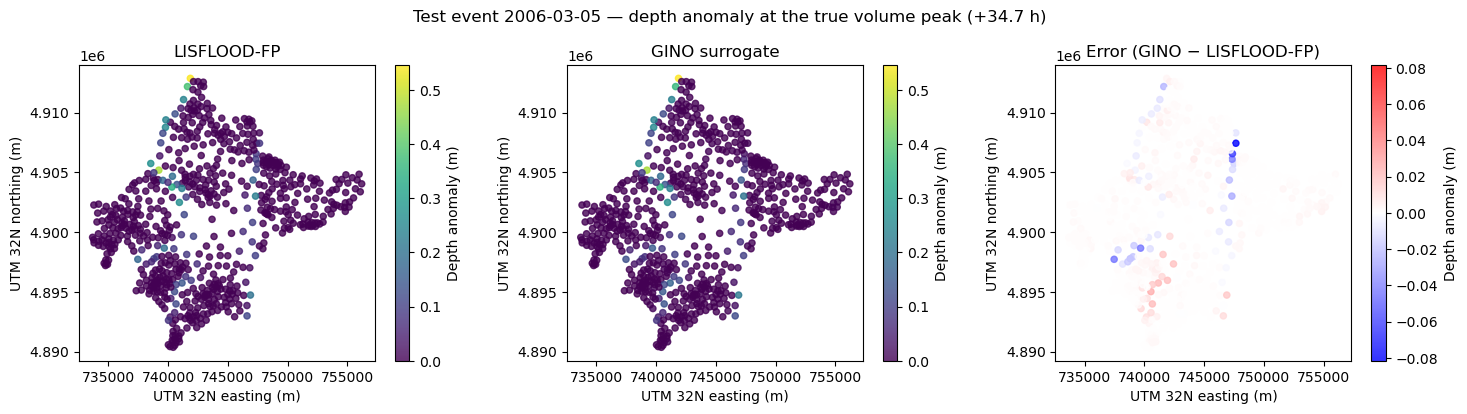}
  \caption{Node depth anomaly (m) at the time of the true volume peak (\SI{34.7}{h}) for the test event 2006-03-05: LISFLOOD-FP (left), GINO surrogate (centre) and error, prediction minus truth (right; red = overestimation, blue = underestimation). Each dot is a node of the 659-node Voronoi mesh; coordinates in UTM~32N (m).}
  \label{fig:peakmaps}
\end{figure*}

Figure~\ref{fig:overview} shows two test events of different magnitude in detail. For the test event of 2006-03-05 the surrogate tracks the inundated fraction of the municipality, the flood volume and the depth anomaly at the most affected node almost exactly through the rising limb and the peak, including the two-stage rise of the node hydrograph; in the recession it slightly overestimates the remaining volume and wet area. For the much smaller event of 2016-03-18 the flood volume is again reproduced closely, with a second, slightly overestimated peak around \SI{55}{h}; the very small wet area (below 2\% of the evaluated area) is underestimated, and at the most affected node the timing of the rise is correct but its amplitude is underestimated (about \SI{0.10}{m} against \SI{0.14}{m}).

The spatial pattern at the peak of the 2006-03-05 event is shown in Fig.~\ref{fig:peakmaps}. The predicted map reproduces the location and depth of the flooded nodes, from the deepest nodes at the northern tip of the municipality to the flooded nodes west and south of the city. Errors are within a few centimetres almost everywhere (at most about \SI{0.08}{m}), with the largest local underestimation at a few nodes along the eastern river corridor and at the south-western edge of the city.

The most challenging test cases are the two largest floods of the test set, 2019-11-17 and 2024-09-18, with CSI at \SI{0.05}{m} of $0.808$ and $0.820$, respectively. For 2019-11-17 the RMSE is \SI{0.020}{m} and the maximum-depth map still reaches $R^2 = 0.971$; 2024-09-18 is the most demanding event overall (RMSE \SI{0.054}{m}, $R^2 = 0.899$ for maximum depth). 

\subsection{Ensemble forecasting}\label{sec:res-ensemble}

\begin{figure*}[!tb]
  \centering
  \includegraphics[width=\textwidth]{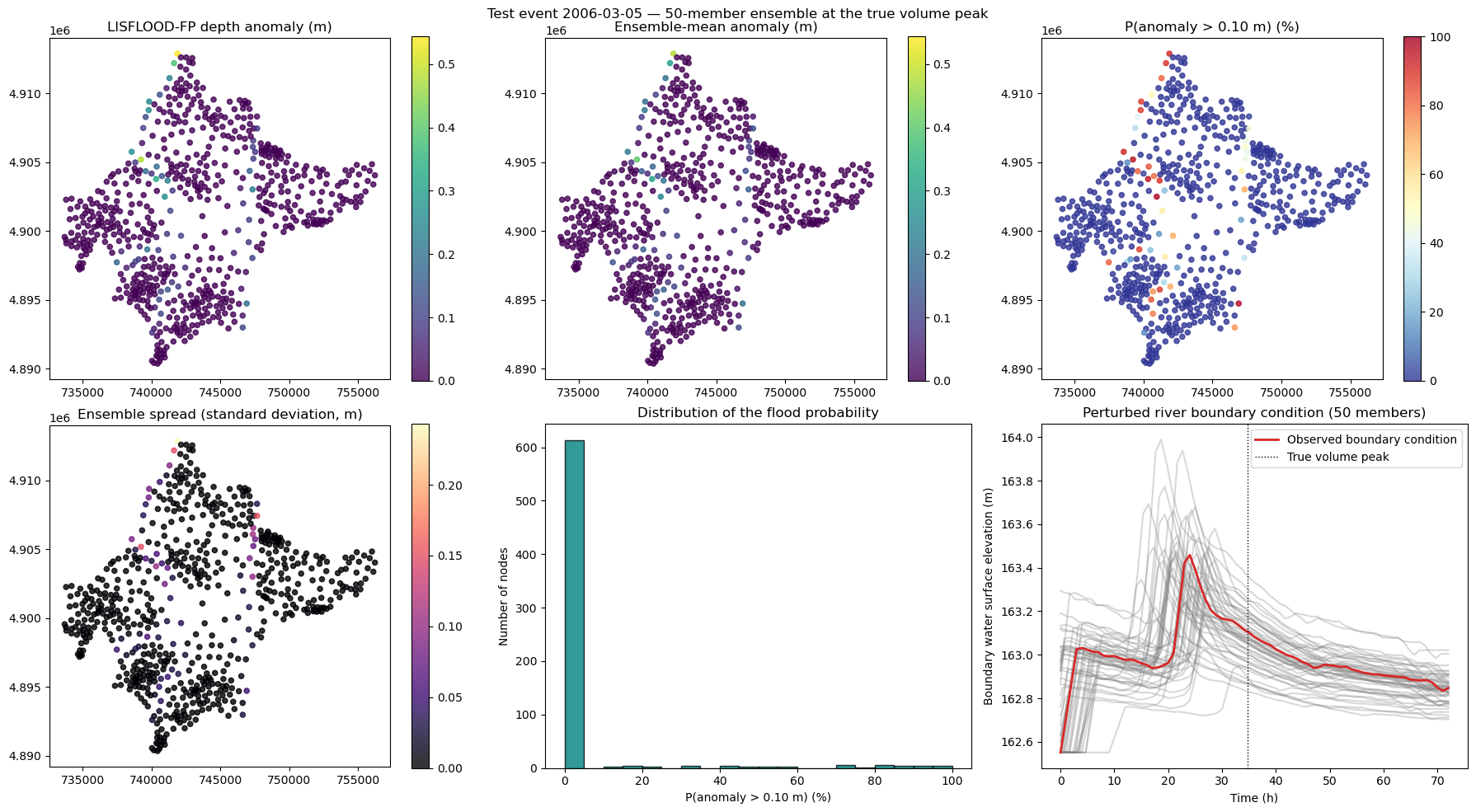}
  \caption{50-member ensemble forecast with the GINO surrogate for a test event, at the time of the true volume peak. Top: LISFLOOD-FP depth anomaly (m), ensemble-mean anomaly (m) and probability of an anomaly above \SI{0.10}{m} (\%). Bottom: ensemble spread (standard deviation, m); histogram of the flood probability over the evaluated nodes; perturbed boundary water surface elevation of the 50 members (grey) and observed boundary condition (red) at the node with the largest stage range, with the time of the true peak (dotted line).}
  \label{fig:ensemble}
\end{figure*}

Because one rollout takes a single batched forward pass per time step, the surrogate can turn the uncertainty of the forcing into a probabilistic flood forecast. For each test event we generated 50 members by perturbing the amplitude and timing of the river hydrographs and of the rainfall (Section~\ref{sec:ensemble}) and ran them in one batched rollout. Figure~\ref{fig:ensemble} shows the result for the 2006-03-05 event. The ensemble mean reproduces the true peak map, and the ensemble spread and the probability of an anomaly above \SI{0.10}{m} concentrate on the flood-prone nodes near the rivers and in the north of the municipality, while the rest of the domain is confidently dry. Over all test events, the mean spread on wet nodes is \SI{0.082}{m} and the median ratio between ensemble spread and error of the ensemble mean is $0.72$, i.e.\ the ensemble is moderately under-dispersive but of the right order of magnitude (a ratio of 1 indicates a consistent ensemble). Only 8.0\% of the nodes have a flood probability between 5\% and 95\%: the forecast is sharp, and the uncertainty is confined to a small set of nodes where it is informative for decision makers.

\subsection{Computational efficiency}\label{sec:res-efficiency}

We compare the wall-clock time needed to produce a \SI{72}{h} forecast for the municipality on the same hardware, an AWS \texttt{g5.2xlarge} instance with one NVIDIA A10G GPU (\SI{24}{GB}). LISFLOOD-FP must simulate its whole computational domain, which is required to resolve the flows entering and leaving the municipality, whereas the surrogate predicts directly the 659 polygon depths of the risk product. For LISFLOOD-FP we report the duration of the time-stepping loop printed by the solver (\texttt{loop time}, snapshot output included); for the surrogate, the end-to-end rollout time, including the construction of the forcing channels, with data already in memory and after one warm-up run.

Over the 254 event simulations of the dataset, a \SI{72}{h} LISFLOOD-FP run takes a median of \SI{55}{s} (range 51--\SI{76}{s}; \SI{57}{s} for the 38 test events). The GINO surrogate produces the same \SI{72}{h} forecast, 216 autoregressive steps, in \SI{1.72}{s}, about 32 times faster. The gain is much larger for ensembles: the 50-member ensemble of Section~\ref{sec:res-ensemble} is computed in a single batched rollout in \SI{6.19}{s}, about \SI{0.12}{s} per member, whereas 50 LISFLOOD-FP runs on the same GPU would take about \SI{46}{min} (\num{50}~$\times$~\SI{55}{s}), a speed-up of about 440.

%% file: sections/05_discussion.tex
\section{Discussion}\label{sec:discussion}

\subsection{Accuracy on independent observed events}
The main result of this study is that a time-dependent GINO keeps high accuracy when it is tested on observed
events that are independent of the training set. In fact, none of the 38 test events, nor any
scaled variant of them, was used for training or model selection. Their hydrograph shapes, rain patterns and
timing therefore differ from those of the 24 training events. Accuracy on the test set is nonetheless close to
the validation accuracy. CSI at \SI{0.05}{m} is $0.934 \pm 0.039$ on the test set against
$0.936 \pm 0.045$ on validation, and the $R^2$ of the maximum depth is $0.984 \pm 0.017$ against
$0.988 \pm 0.014$ (Table~\ref{tab:ffmetrics}). The drop is larger for the deeper threshold (CSI at \SI{0.3}{m}
$0.844 \pm 0.084$ against $0.901 \pm 0.081$) and for the timing metrics, which are the hardest to transfer to
new events. The surrogate also orders events correctly by magnitude: the correlation between true and predicted
peak flood volume across test events is 0.995. This is the information an early-warning system needs to tell a
minor event from a major one.

The node-level metrics confirm this picture (Section~\ref{sec:res-node}). Over the 38 test events, the surrogate
captures the flooded area at the volume peak with a CSI of 0.840, reproduces the peak flood volume with a mean
ratio of 0.971 and has a wet-node MAE of \SI{0.020}{m} over the full \SI{72}{h}. Since none of these events, nor
any variant of them, was available in training, the surrogate does more than reproduce known scenarios: it
composes the response to a new combination of river and rain forcing.

\subsection{Operational value of stage-driven forcing}
Driving the surrogate with river stage instead of discharge is a deliberate design choice. Stage is what the
regional network measures and transmits in real time \citep{arpae_dext3r}. Converting it to discharge would add
a rating-curve error that is largest exactly during floods, when the curves are extrapolated and the channel
geometry may change. The first-downstream-gauge rule ties each inflow to a single, well-defined observation, and
the library-anchored calibration of the stage boundary conditions changed the per-river scales by moderate
factors (Section~\ref{sec:lisflood}). As a result, the full chain---observed gauges, boundary conditions,
LISFLOOD-FP, surrogate---can be run in the same way in hindcast and in forecast mode, using stage forecasts or
observed stages as they arrive. The same pipeline also produced the test set: 38 of the 43 candidate events passed
the automatic quality control of the driver gauges.

\subsection{Product-oriented mesh and volume-conserving targets}
The Voronoi mesh was designed for the risk product, with seeds placed according to exposure (buildings,
transport, geology, water), not for the hydraulics. Predicting directly on it means that the surrogate output is
already in the units used to report risk per area, with no post-processing. Polygon-mean depths conserve the
flood volume of the LISFLOOD-FP solution by construction. Because the number of nodes and the architecture are unchanged,
this choice does not add to the inference cost. On the training side, weighting the loss on predicted wet nodes
and extending the push-forward curriculum to 32 unrolled steps target spurious thin water and long-horizon drift,
the two typical failure modes of autoregressive surrogates \citep{brandstetter2022mpnn,taghizadeh2025hydrographnet};
both act only on training and leave the inference cost unchanged.

\subsection{Ensembles and simulator credibility}
Because all ensemble members share the static inputs and the mesh, a 50-member ensemble is computed as a single
batched rollout per event (Section~\ref{sec:res-ensemble}). The ensemble produces flood-probability maps in which
8.0\% of the nodes carry informative probabilities (between 5\% and 95\%). These nodes mark the uncertain fringe
of the flood, which is where forecasters should focus. The median spread-to-error ratio of 0.72 shows that the
current perturbations of river stage and rain are somewhat under-dispersive. Calibrating the perturbation
amplitudes against forecast errors of the stage and rain predictions is a natural next step.

Finally, a surrogate can only be as good as its simulator. Surrogate studies usually treat the hydraulic
simulations as ground truth; we additionally compared the LISFLOOD-FP set-up with the regional
perimeter of the May 2023 flooding \citep{rer2025perimetrazione} and obtained a CSI of 0.6
(Section~\ref{sec:res-simulator}). We used the regional perimeter rather than the Copernicus EMS radar delineation
\citep{cems2023emsr664}, because the radar misses much of the flooding in built-up areas. Binary skill scores
depend strongly on flood magnitude, resolution and the reference data \citep{stephens2014binary,wing2017validation}.
For a \SI{100}{m} model without subgrid channels or levee and drainage detail, compared with a 1:5000
photo-interpreted envelope, this agreement indicates that the simulator captures the main flooded areas of the
largest recent event. It thus provides a credible physical basis for the surrogate.

\subsection{Limitations and future work}\label{sec:limitations}
Several limitations point to future work.
(i)~The surrogate was trained and tested on a single municipality. Training on several cities is the natural route towards a future implementation.
(ii)~The surrogate inherits the limitations of its simulator: a \SI{100}{m} LISFLOOD-FP set-up without subgrid channels, in which levees are represented only as raised DEM cells at crest elevation (overtopping only, no breaching), and without the urban drainage network, which limits the detail of urban flow paths. 
(iii)~Results are from a single training seed.

%% file: sections/06_conclusions.tex
\section{Conclusions}\label{sec:conclusions}

This study presented a time-dependent geometry-informed neural operator surrogate of LISFLOOD-FP for rapid
pluvial--fluvial flood forecasting in the municipality of Forl\`{i}, Italy. The surrogate is driven only by the
observations available in real time: river stage and rain gauges. Building on the time-dependent GINO
formulation of \citet{taghizadeh2026floodforecaster}, we developed a reproducible pipeline that turns
public regional observations into a physically based training set. In this pipeline each river inflow is driven
by its first downstream stage gauge through calibrated boundary conditions, and rain enters as SCS-CN effective
rainfall. An autoregressive GINO, trained on 216 simulations of 24 observed events with scaled forcing, predicts
volume-conserving polygon-mean depths on an exposure-based Voronoi mesh of 659 nodes every \SI{20}{min} over a
\SI{72}{h} horizon.

Tested on 38 independent observed events that were never seen in training, the surrogate reaches a CSI of 0.934 at
\SI{0.05}{m} and 0.844 at \SI{0.3}{m}, a depth RMSE of \SI{0.008}{m} and an $R^2$ of 0.984 for the maximum-depth
maps. These values are close to its accuracy on the validation simulations. At node level, it captures the
flooded area at the volume peak with a CSI of 0.840 and the peak volume with a mean ratio of 0.971.
Volume-conserving targets preserve the flood volume of the simulator on
the product mesh at no additional inference cost. On the same GPU, a 72-h forecast takes \SI{1.7}{s}
against \SI{55}{s} for LISFLOOD-FP, and a 50-member ensemble, computed in one batched rollout, takes
\SI{6.2}{s} against about \SI{46}{min}, and produces flood-probability maps for early warning. The underlying
LISFLOOD-FP set-up reproduces the observed May 2023 flood extent with a CSI of 0.6.

Together, these results show that neural-operator surrogates can move from controlled experiments on synthetic scenarios to an
operational setting: driven by stage rather than discharge, forced by both rivers and rain, delivered on the units
of the risk product, and tested on real events. Future work will extend the approach to several cities through
multi-domain training and domain adaptation, refine the mesh near the rivers, enrich the training set with more
extreme events, and couple the surrogate with operational stage and rain forecasts. These steps lead towards fast
ensemble-based flood early warning for the Emilia-Romagna region.
